\documentclass{llncs}
\usepackage[T1]{fontenc}
\usepackage{graphicx}
\usepackage{amsmath}
\usepackage{amsfonts}
\usepackage{dsfont}
\usepackage{algorithm}
\usepackage{algorithmic}
\usepackage{subfigure}
\usepackage{wrapfig}
\usepackage{multirow}
\usepackage{adjustbox}
\usepackage{fancyhdr}
\begin{document}
\title{Sample Efficient Robot Learning in Supervised Effect Prediction Tasks}
%
%
\author{Mehmet Arda Eren\inst{1}\orcidID{0009-0009-0259-8441} \and
Erhan Oztop\inst{1,2}\orcidID{0000-0002-3051-6038}}
%
\institute{Department of Artificial Intelligence, Ozyegin University, Istanbul, Turkey \and
OTRI/SISReC, Osaka University, Osaka, Japan\\
\email{arda.eren@ozu.edu.tr}\\
\email{erhan.oztop@\{otri.osaka-u.ac.jp, ozyegin.edu.tr\}}
}
\maketitle              
\begin{abstract}
In self-supervised robotic learning, agents acquire data through active interaction with their environment, incurring costs such as energy use, human oversight, and experimental time. To mitigate these, sample-efficient exploration is essential. While intrinsic motivation (IM) methods like learning progress (LP) are widely used in robotics, and active learning (AL) is well established for classification in machine learning, few frameworks address continuous, high-dimensional regression tasks typical of world model learning.
We propose MUSEL (Model Uncertainty for Sample-Efficient Learning), a novel AL framework tailored for regression tasks in robotics, such as action-effect prediction. MUSEL introduces a model uncertainty metric that combines total predictive uncertainty, learning progress, and input diversity to guide data acquisition. We validate our approach using a Stochastic Variational Deep Kernel Learning (SVDKL) model in two robotic tabletop tasks. Experimental results demonstrate that MUSEL improves both learning accuracy and sample efficiency, validating its effectiveness in learning action effects and selecting informative samples.

\end{abstract}


\section{Introduction}
A central challenge in robotics is to equip them with planning and control mechanisms for task execution in complex environments. Given an action repertoire, learning internal models for action-effect relations through interaction with the environment is a promising approach \cite{ahmetoglu2022deepsym,seker2019deep}. In simulation, this often involves executing random actions and recording outcomes, which are then learned in supervised fashion akin to constructing a simplified world model \cite{ha2018world}. However, unguided action selection severely limits sample efficiency, rendering such approaches impractical for real-world deployment due to high execution costs. While sample efficiency is often explored in the context of reinforcement learning (RL) \cite{baranes2013active,li2023ocba}, it is relatively less addressed in self-supervised learning \cite{oudeyer2007intrinsic,pere2018unsupervised,bugur2019effect}--a gap our work aims to fill. Intrinsic motivation (IM) and active learning (AL)-based approaches may help address this if they can be adopted to robot learning. IM drives agents to explore their environment based on learning progress or novelty, encouraging discovery without explicit rewards. However, it does not directly target sample efficiency. On the other hand, AL improves efficiency by selecting the most informative data points, reducing the need for extensive labeling in classification tasks. AL-based methods guide data selection using three key measures \cite{settles2009active}: \emph{informativeness} \cite{liu2021pool,wu2018pool}, \emph{representativeness} \cite{wu2018pool}, and \emph{diversity} \cite{liu2021pool,wu2018pool,wu2019active}. However, applying these to self-supervised robotic tasks in continuous action and effect spaces is challenging, as they typically assume a finite data pool or require approximations of the state-action space \cite{taylor2021active}, which are computationally expensive to compute \cite{van2020population}.
In this study we propose a novel active learning algorithm, \textbf{Model Uncertainty for Sample-Efficient Learning (MUSEL)}, and show its application on the self-supervised learning of effect prediction tasks in continuous action spaces. MUSEL uses a Stochastic Variational Deep Kernel Process (SVDKL) \cite{wilson2016stochastic} to jointly estimate data-and-model uncertainty and track learning progress (LP) \cite{oudeyer2007intrinsic}. Combined with max-min distance between input samples \cite{wu2019active}, these signals approximate model uncertainty and work together to improve sample efficiency.
To evaluate MUSEL, we test it in a simulated environment where a 7-DOF robot interacts with rigid body objects. Experiments are conducted in two distinct scenarios--one-sphere and two-sphere setups--to assess the model's sample efficiency. Ablation studies further dissect MUSEL’s components, revealing their individual and combined contributions to the performance. The main contributions of this study can be summarized as follows:
\begin{itemize}
    \item Development of a concrete active learning algorithm for action-effect prediction applicable to continuous action and state settings.
    \item Extraction of model uncertainty from total uncertainty using learning progress and input diversity.
    \item Use of model uncertainty-based sample selection in robot action effect prediction learning.
    \item Demonstration of the applicability of our method to nontrivial robot self-supervised learning tasks.
\end{itemize}
\section{Related Work}\label{sec:related-work}
\subsection{Active Learning and Its Application in Robotics} \label{subsec:al_robotics}
Active learning (AL) \cite{cohn1996active}, characterized by the strategic selection of training data, enhances model performance while reducing costs and has been applied across diverse domains. AL algorithms mainly address supervised tasks, developed for both classification \cite{fu2013active,liu2021pool} and regression \cite{wu2018pool,wu2019active}. These methods can be formally categorized based on the presence of an iterative process: supervised \cite{fu2013active,wu2018pool,wu2019active} if it exists, unsupervised \cite{liu2021pool} otherwise. Another categorization is based on the nature of the sampling space: pool-based methods use a finite set, while population-based methods operate over an infinite set \cite{sugiyama2009pool}. To address the continuous input-output characteristics typical of robotic tasks, MUSEL is proposed as a supervised, population-based AL framework for regression.

AL has been adopted in diverse robotic domains, including environmental mapping, nonparametric shape estimation, control, perception and prediction, and distributed learning \cite{lluvia2021active,maeda2017active,taylor2021active}.
However, these methods often treat exploration and efficiency as separate concerns. Our approach addresses this by utilizing uncertainty-based sampling to balance the two, thereby enabling sample-efficient supervised effect prediction.
In addition, intrinsic motivation (IM) improves sampling efficiency by selecting actions based on different internal signals (\cite{houthooft2016vime,marshall2004emergent,oudeyer2007intrinsic}). Among these methods, learning progress (LP) motivation--effective in robotic settings \cite{bugur2019effect,say2023model}--aids our sample selection by estimating data uncertainty and is evaluated against MUSEL.
From a neuroscience perspective, the underlying principle of MUSEL can be viewed as adjusting system parameters to reduce uncertainty, which aligns with the term \emph{``Active Sensing''} \cite{yang2016theoretical}, where actions aim to maximize information gain. 
Other applications of this method focus on task relevance \cite{greigarn2019task} or trajectory optimization \cite{chen2021computationally}, whereas we focus on using model uncertainty for sample-efficient learning in continuous spaces.
\subsection{Uncertainty Quantification} \label{subsec:uq}
Uncertainty refers to the expected error or confidence level of a model’s predictions. It is a central metric in active learning for sample selection \cite{yang2016active,yang2015multi}, typically requiring models to output both predictions and corresponding uncertainties \cite{gawlikowski2021survey}. Approaches have been developed for both classification \cite{mena2021survey,sensoy2018evidential} and regression \cite{gawlikowski2021survey,tripathy2016gaussian} tasks.
Gaussian Process Regression (GPR) is a common choice for regression, providing a Normal distribution whose mean and standard deviation represent the prediction and its uncertainty, correspondingly \cite{williams1995gaussian}. However, its scalability is limited due to increased computational demands with larger datasets \cite{tripathy2016gaussian}.
Stochastic Variational Gaussian Processes (SVGP) \cite{hensman2013gaussian} overcome this by introducing a sparse, variationally optimized inducing point framework, enabling true stochastic (mini-batch) training via the Evidence Lower Bound (ELBO) \cite{blei2017variational}. Jointly optimizing $m \ll n$ inducing points, where $n$ is the number of samples, reduces inference complexity from $\mathcal{O}(n^3)$ to $\mathcal{O}(nm^2)$ for fitting and $\mathcal{O}(m^2)$ per-point prediction, allowing SVGP to scale to large or streaming datasets.

Likewise, to enable Gaussian Processes to capture complex, high-dimensional patterns, DKL \cite{wilson2016deep} uses a neural network to learn feature representations before applying the GP kernel, thus combining expressive deep features with the non-parametric flexibility of GPs.
Stochastic Variational Deep Kernel Learning (SVDKL) \cite{wilson2016stochastic} further builds on this by applying an SVGP to the learned features, with joint optimization of both the neural network and the GP via a mini-batch ELBO. This setup supports scalable training on large datasets while maintaining well-calibrated uncertainty estimates. In this work, we adopt SVDKL as the backbone of our learning framework.


\section{Method}
\label{sec:method}

\subsection{Problem Domain Formalization}

We consider a robot tasked with learning the \emph{effects} of its \emph{actions} through interaction with its environment. This is achieved by collecting experience tuples of the form \emph{(state, action)} $\rightarrow$ \emph{effect}, where the \emph{state} describes the configuration of both the robot and the environment. While we present our method in the context of a specific robotic task environment used for evaluation, the approach is generalizable to any effect prediction setting. For prediction, we represent the input, i.e., state-action pair, as $x=(s,a)$. Thus input set is defined as 
\[
\mathds{X} = \left\{\{x_t=(s_t, a_t)\}_{t=1}^T, \  s_t \in S \subset \mathbb{R}^m, a_t \in A \subset \mathbb{R}^n \right\},
\]
where $t$ is the sample step, $T$ is the total number of steps executed, \(S\) and \(A\) denote the state and action spaces, respectively, and \(m\) and \(n\) are their corresponding dimensionalities. Here, \(T\) grows as the robot interacts with the environment. In this work, the state \(s\) is a 2D vector representing coordinates of the target sphere, and the action \(a\) is a real-valued scalar indicating hit angle. The robot is allowed to specify both the initial position of the sphere within a predefined workspace and the angle at which it will strike. The resulting effect is defined as the displacement in the 2D space after the execution of the action. Thus set of effects is given by
\[
\mathds{Y} = \{s'_t \mid s'_t = f(s_t, a_t),\ s'_t \in \mathbb{R}^r, \ (s_t, a_t) \in \mathds{X}\}_{t=1}^T
\]
where $f$ is the environment transition function, \(r\) denotes the dimensionality of the effect space.
\subsection{Proposed Method: MUSEL}
\begin{figure}[!htb]
    \centering
    \includegraphics[width=\linewidth]{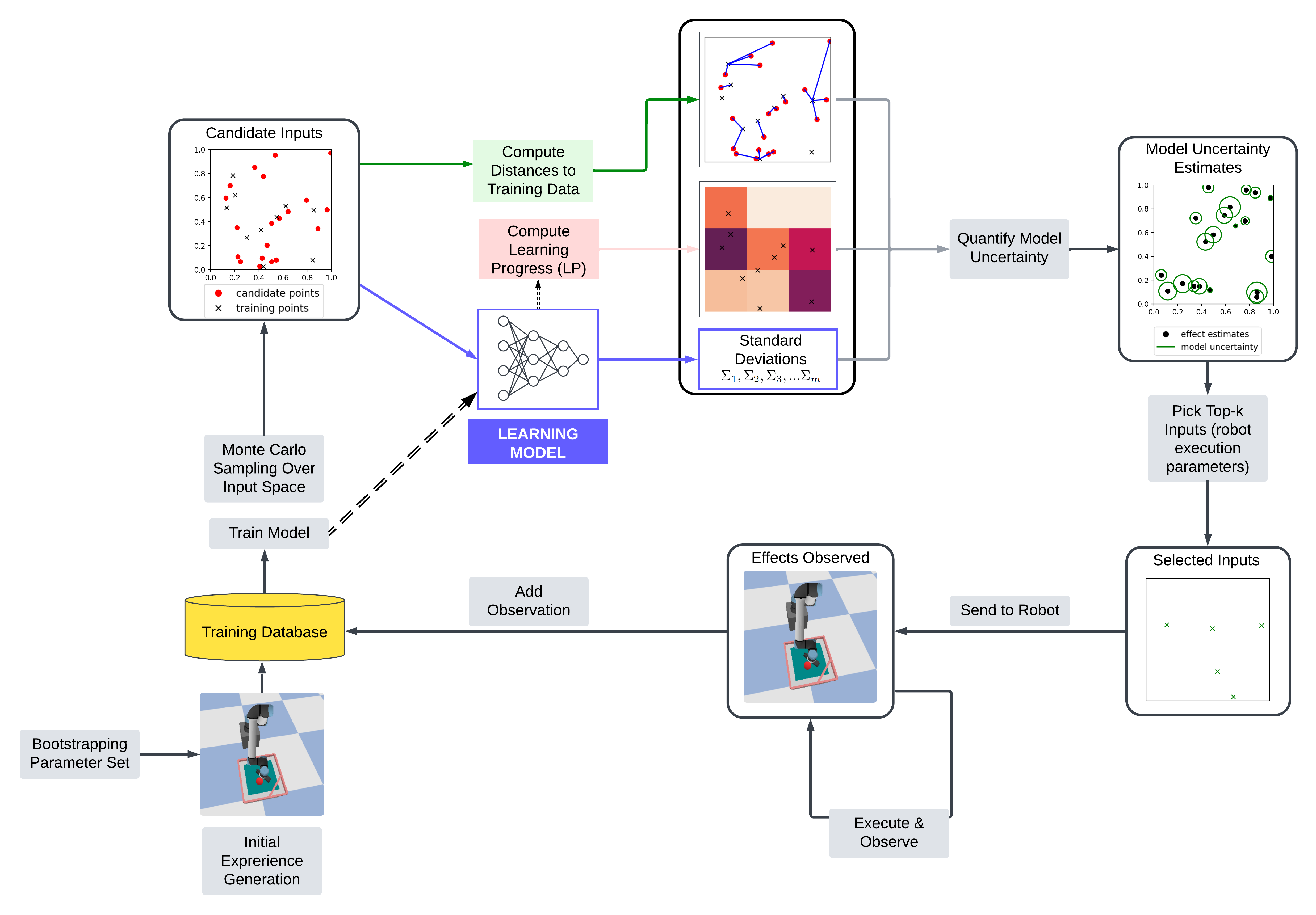}
    \caption{The proposed model, MUSEL is illustrated. See text for details.}
    \label{fig:MUSEL_diagram}
\end{figure}
In this section, we explain our proposed model, MUSEL, which aims to minimize the number of observation queries while learning an input-output mapping over continuous spaces without sacrificing prediction accuracy. At the core of MUSEL is the model uncertainty estimation method introduced in Section~\ref{subsec:estimating_model_uncertainty}, which is embedded in a loop comprising \textit{execution sample selection, robot execution, model learning, and input sampling}, as illustrated in Figure~\ref{fig:MUSEL_diagram}.
Learning with MUSEL is bootstrapped by an initial learning effectively initializing the AL loop. The initial dataset composes of $m_{init}\geq1$ random action execution experiences, and is used to train the SVDKL model (see Section~\ref{subsec:svdkl}). Thereafter, a candidate set of state–action pairs is generated, and model uncertainties for these candidates are estimated as described in Section~\ref{subsec:estimating_model_uncertainty}. The state–action pairs with the top $k$ uncertainty values are executed by the robot and included in the training dataset with their observed effects. The loop then returns to the model training phase as outlined in Section~\ref{subsec:algo}.

\subsection{SVDKL Backbone} \label{subsec:svdkl}

As reviewed in Section~\ref{subsec:uq}, Stochastic Variational Deep Kernel Learning (SVDKL) uses a trainable neural network to map inputs into a rich feature space and an SVGP with a small set of inducing points to perform regression.
Let $h_\phi: \mathbb{R}^{m+n}\to\mathbb{R}^d$ be the neural network parameterized by $\phi$. This network acts as a feature extractor by mapping each input $x_t=(s_t,a_t)$ to a $d$-dimensional feature $z_t = h_\phi(x_t)$. An SVGP model is then placed on top of these features \cite{wilson2016stochastic}, containing $M$ inducing inputs $Z=\{z_j\}_{j=1}^M$ (where $M\ll N$, and $N$ is the number of samples) and corresponding latent outputs $u=f(Z)$. Since the true posterior $p(u\mid \{z_i,y_i\}_{i=1}^N)$ is intractable, we approximate it via variational inference by defining:
\begin{equation}
    q(u) = \mathcal{N}(u \,;\, m,\, S).
\end{equation}
Here, $m \in \mathbb{R}^M$ and $S \in \mathbb{R}^{M \times M}$ represent the mean and covariance of the variational distribution $q(u)$ \cite{hensman2013gaussian}. Thus, training and inference are conducted using information summarized by these $M$ inducing points, rather than directly on the full dataset. We maximize the mini-batch ELBO to ensure that $q(u)$ explains the observed data while remaining close to the prior $p(u)$ \cite{hensman2013gaussian}:
\begin{equation}
    \mathcal{L}(\phi, m, S) \;=\; 
    \sum_{i\in\mathcal{B}} \mathbb{E}_{q(u)}\bigl[\log p(y_i \mid f(z_i))\bigr]
    \;-\;
    \mathrm{KL}\bigl[q(u)\,\|\,p(u)\bigr],
\end{equation}
where $\mathcal{B}$ denotes a random subset of training indices. Monte Carlo sampling through the GP head and stochastic gradient descent on $(\phi, m, S)$ enable end-to-end learning of both feature and GP parameters at scale \cite{wilson2016stochastic}. This SVDKL backbone yields calibrated predictive means and variances for any input $x$, which are utilized for uncertainty-driven sample selection in the following section. Details of the model are provided in Section~\ref{subsec:overview}.

\subsection{Estimating Model Uncertainty}
\label{subsec:estimating_model_uncertainty}

The total uncertainty associated with a prediction is a compound of data and model uncertainties \cite{gawlikowski2021survey}. We take the \emph{std} output of SVDKL for input $x$ (i.e., a state-action pair), denoted $\sigma(x)$, as the total uncertainty $U(x)$. During the data collection process, only model uncertainty can be reduced by acquiring additional samples; therefore, our objective is to estimate this uncertainty to effectively guide sample selection. Following \cite{lee2024data}, we assume a statistical independence between the model uncertainty, $U_{model}$ and data uncertainty, $U_{data}$ and hence write
\begin{equation} \label{eq:unc_assumption}
    U(x) = \sigma(x) = U_{model}(x)U_{data}(x).
\end{equation}
Since $\sigma(x)$ is predicted by SVDKL, if $U_{data}(x)$ can be estimated, then $U_{model}(x)$ can also be inferred. As LP quantifies the decline in learning loss, a low LP indicates saturated learning and suggests high $U_{data}$. Likewise, regions where dense training data coincide with high total uncertainty, reflecting persistent error, indicate raised $U_{data}$. Based on the inverse relationship between these effects, we propose the following data uncertainty estimate:
\begin{equation} \label{eq:data_unc}
    U_{data}(x) = \frac{1}{\ell(x)LP(x)},
\end{equation}
where $\ell(x)$ denotes the minimum distance (MD) to the training inputs. Note that the maximum of this value is often used in active learning algorithms for sample selection \cite{wu2019active}. Specifically, the algorithm evaluates the minimum distance to the set of training inputs $O$, given by:
\begin{equation} \label{eq:min_dist}
    \ell(x) = \min\{||x - x_k||\} , \text{ where } x_k \in O.
\end{equation}
To compute $U_{model}(x)$ numerically, we still need to estimate the LP associated with input $x$. Since the input space is continuous, computing the LP for every individual input is infeasible. To address this, we assume locality and assign a single LP value to all points within each region by partitioning the input space into equally spaced $n$-dimensional boxes.
Last $p$ average RMSE errors for each region $r_i$ are kept ($E^t(r_i)$),  and are used to make a linear fit against $t=1,2,..,p$ so that $E^t(r_i) \approx m_it + c_i$. The average errors are calculated with 
\begin{equation}
    E^t(r_i) = \frac{1}{|r_i|}\sum_{x' \in r_i} e(x'), 
\end{equation}
where $e(x')$ indicates the prediction error on the executed input $x'$ that falls in the region $r_i$. With the following definition, LP is mapped to $[10^{-4}, 1]$, so that it reflects the (normalized) error reduction rate when the error is decreasing; otherwise, it is set to a small positive value to ensure mathematical consistency in Equation~\ref{eq:model_unc}.
\begin{equation} \label{eq:lp}
    \underset{x \in r_i}{LP(x)} = LP(r_i)=
    \begin{cases}
        -\frac{2}{\pi}arctan{(m_i)},& \text{if } m_i\leq 0\\
        10^{-4},              & \text{otherwise}
    \end{cases}
\end{equation}
Substituting $U_{data}$ in Equation~\ref{eq:unc_assumption} with Equation~\ref{eq:data_unc}, model uncertainty becomes
\begin{equation} \label{eq:model_unc}
    U_{model}(x) = \sigma(x)  \ell(x)  LP(x).
\end{equation}
\subsection{Sample Selection Based on Model Uncertainty} \label{subsec:algo}
In this section, we describe the details of how our model, MUSEL, selects samples based on the computational elements discussed in the previous sections. To address the continuous nature of action-effect prediction tasks, a population-based active learning approach is used, approximating the candidate set via Monte Carlo sampling.

\begin{algorithm}[!htb]              
  \caption{Active Learning with MUSEL}\label{algo:al}
  \algsetup{indent=1em}             
  \begin{algorithmic}[1]            
    \REQUIRE environment $Env$, $n_{iter}$, $m_{init}$, $m_{cand}$, $k$
    \ENSURE model $M$, dataset $\mathds{D}_O$
    \STATE $\mathds{X}_O \leftarrow \textsc{CreateSet}(m_{init})$
      \COMMENT{Initialize the training dataset}
    \STATE $\mathds{Y}_O \leftarrow \textbf{ROBOT}:\textsc{ExecuteAndObserve}(\mathds{X}_O)$
    \STATE $M \leftarrow \textbf{MODEL}:\textsc{Initialize}$
    \FOR{$i\gets 1$ \TO $n_{iter}$}
      \STATE $M \leftarrow \textbf{MODEL}:\textsc{Train}(\mathds{X}_O,\mathds{Y}_O)$
      \STATE $\mathds{X}_C \leftarrow \textsc{SampleInputSpace}(m_{cand})$
        \COMMENT{(Re)create candidate inputs}
      \STATE $U_{\text{model}} \leftarrow \textsc{EstimateModelUncertainty}(\mathds{X}_C,\textbf{MODEL})$
      \STATE $X^*_C \leftarrow \textsc{SelectTopK}(\mathds{X}_C,U_{\text{model}}, k)$
      \STATE $Y^*_C \leftarrow \textbf{ROBOT}:\textsc{ExecuteAndObserve}(X^*_C)$
      \STATE $\mathds{X}_O \leftarrow \mathds{X}_O \cup X^*_C$
      \STATE $\mathds{Y}_O \leftarrow \mathds{Y}_O \cup Y^*_C$
    \ENDFOR
    \RETURN $M,\mathds{D}_O$
  \end{algorithmic}
\end{algorithm}
The learning begins by generating $m_{iter}$ i.i.d. random state-action parameters to initialize the observation set $\mathds{X}_O$. The robot executes these inputs and records the resulting effects as $\mathds{Y}_O$ (Algorithm~\ref{algo:al}, lines~1-2). Subsequently, a model capable of learning input-output relations with uncertainty estimates, in this case SVDKL, is initialized, and the AL loop starts.
At each iteration, the model is trained on the observed samples, followed by the creation of $\mathds{X}_C$ through random sampling of $m_{cand}$ i.i.d. points from the input space (Algorithm~\ref{algo:al}, lines~5–6). The \textbf{MODEL} and $\mathds{X}_C$ are then used to estimate model uncertainty, as described in Section~\ref{subsec:estimating_model_uncertainty}.
Based on these estimations, candidate inputs are ranked by $U_{\mathrm{model}}$, and the top $k$ are selected as $X^*_C$. These state-action pairs are then executed by the robot, their observed effects are collected ($Y^*_C$), and both are incorporated into the training dataset (Algorithm~\ref{algo:al}, lines~9-11).
The model is then trained with the updated dataset and subsequently, the steps for model uncertainty quantification are carried out. This sampling and learning loop continues until \(n_{iter}\) iterations are completed.


\section{Experiments and Results} \label{sec:experiments}
The environment for the learning experiments is designed as a confined table within the workspace of the robot. The objects can bounce off the boundary walls according to the simulated physics. A diagonal wall spanning two adjacent edges is included to breaks spatial symmetry (see Figure~\ref{fig:tasks}).
\subsection{Experiments Overview} \label{subsec:overview}
\vspace{-4mm}
\begin{figure}[!htbp]
    \label{fig:task_view}
    \centering
    \subfigure{%
    \adjustbox{width=0.28\textwidth,height=0.19\textheight}{%
          \includegraphics{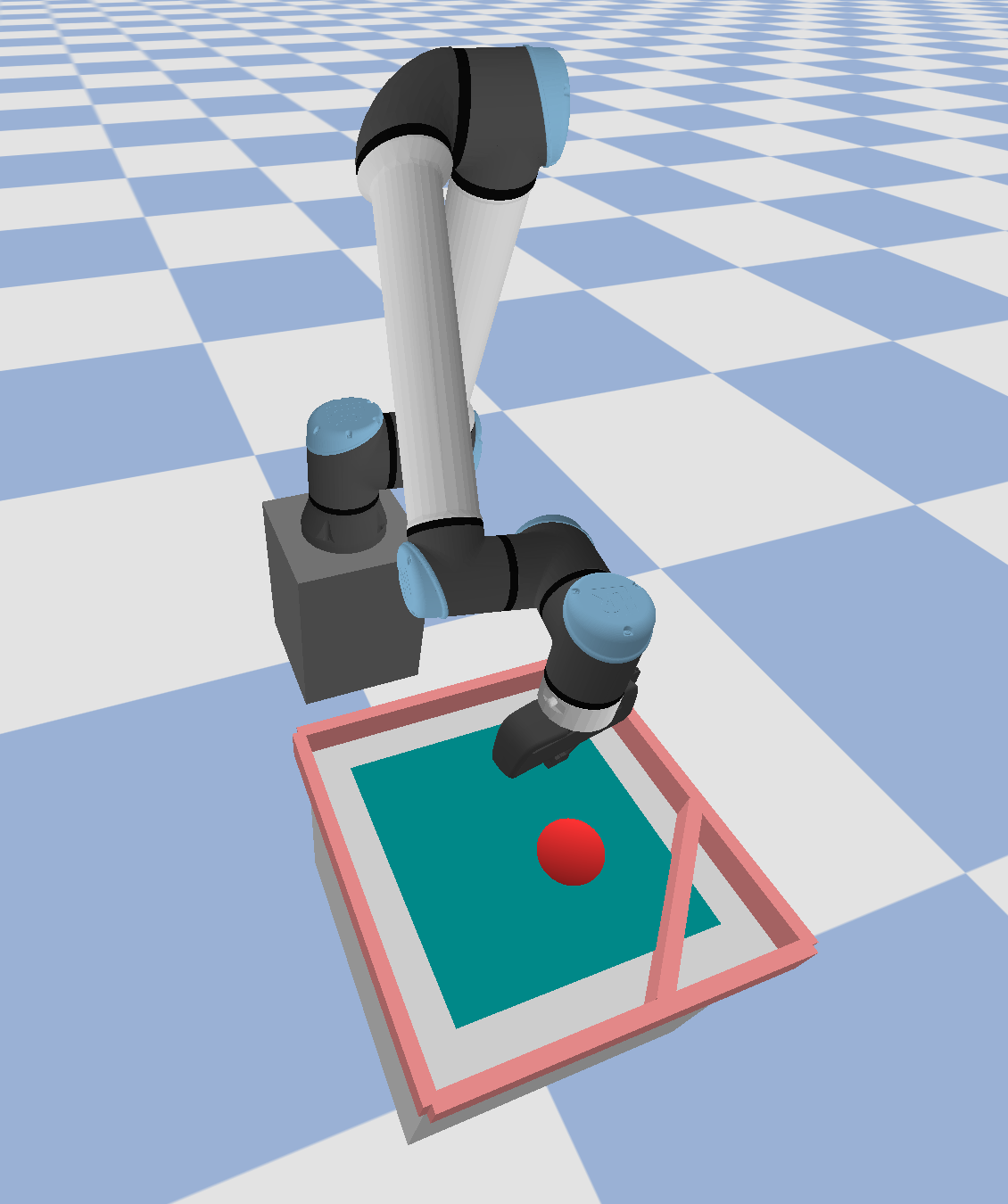}%
        }%
    }%
    \hspace{8em}
    \subfigure{%
    \adjustbox{width=0.28\textwidth,height=0.19\textheight}{%
          \includegraphics{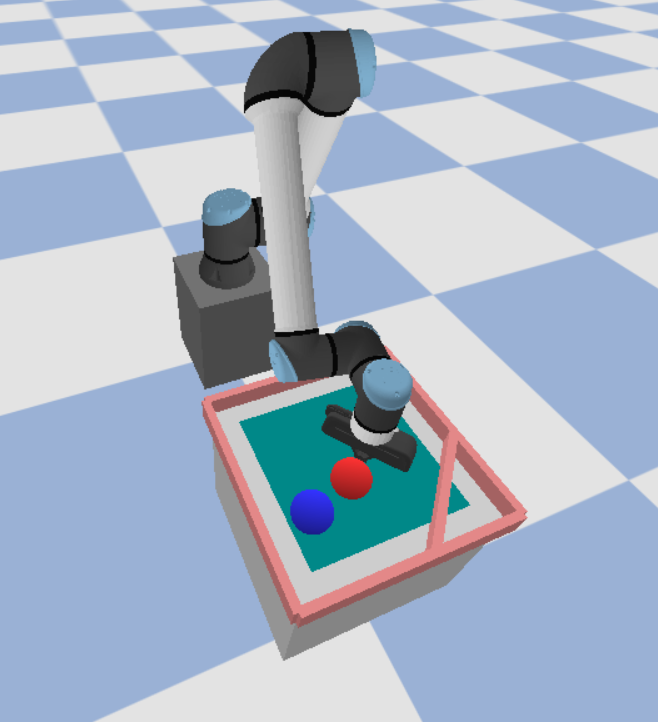}%
        }%
    }%
    \caption{(a) One-sphere setting: The sphere is placed at the location selected by the robot prior to the push action. The blue region indicates allowed positions for the sphere. The diagonal wall is included to increase the task difficulty. (b) Two-sphere setting: This configuration is identical to (a) except for a second sphere, which moves according to physical dynamics but is set to a fixed location before action execution.
    }
     \label{fig:tasks}
\end{figure}
\vspace{-4mm}
%
In the experiments, the robot is equipped with a single action, $push(\alpha)$, where $\alpha \in [-\pi/3, \pi/3] \subset \mathbb{R}$. To execute this action, the robot's end effector first moves to $p_{start}$ and then strikes the object by moving to $p_{end}$, where

\begin{equation} \label{eq:robot_push_calc}
 \begin{gathered}
     p_{start} = (pos_x - r \cos{\alpha}, \: pos_y - r  \sin{\alpha}) \\
     p_{end} = (pos_x + r \cos{\alpha}, \: pos_y + r  \sin{\alpha})
 \end{gathered}
\end{equation}

The robot first records the object's initial position, executes the push action, and then collects its final position once the object comes to rest. The effect, $y = {\delta_x, \delta_y}$, is computed as the difference between the final and initial positions.
This experience is stored as an input-output pair $\{x, y\}$ and added to the training data (Algorithm~{\ref{algo:al}}, lines 10–11), where $x$ defines the robot action and the initial state as $x = \{\sin{\alpha}, \cos{\alpha}, pos_x, pos_y\}$. Note that, as a common practice, $\alpha$ is encoded as $(\sin{\alpha}, \cos{\alpha})$ to maintain input locality over the range of $\alpha$.

In the second experiment, additionally, a fixed‐initial‐position sphere that the robot cannot directly manipulate is introduced (see Fig.~\ref{fig:tasks}b). Its position is excluded from the input and output spaces; it constrains the first sphere’s initial positions to prevent collisions and serves as a dynamic collision object, thereby increasing task complexity.

To implement MUSEL, we set the following meta-parameters: number of active selection iterations $n_{iter}=3000$; initial training dataset size $m_{init}=1$; and, for candidate inputs, $m_{cand}=500$ and $k=1$, representing the numbers of generated and selected inputs, respectively.
LP regions are defined as a $7\times 7\times 7$ grid in the $\alpha \times pos_x \times pos_y$ input space.
In SVDKL, the feature extractor is a three-layer MLP with dimensions $4 \times 32 \times 64 \times 32$, using ReLU activations after every layer except the last; the SVGP part utilizes $M=20$ inducing points and an RBF kernel. We adopt a prioritized training scheme, selecting the $m_{train} = 2000$ least-trained samples and training for one epoch with a learning rate of $5\times10^{-3}$.

Methods were run with ten random seeds and evaluated on test sets uniformly sampled over $25 \times 20 \times 20$ and $20 \times 25 \times 25$ grids for the one-sphere and two-sphere settings, respectively, where the grid dimensions correspond to the push angle and the $x$ and $y$ positions of the sphere.

\subsubsection{LP Values} \label{ssubsec:lp_values}

\begin{figure}[!t]
    \centering
    \includegraphics[width=\textwidth]{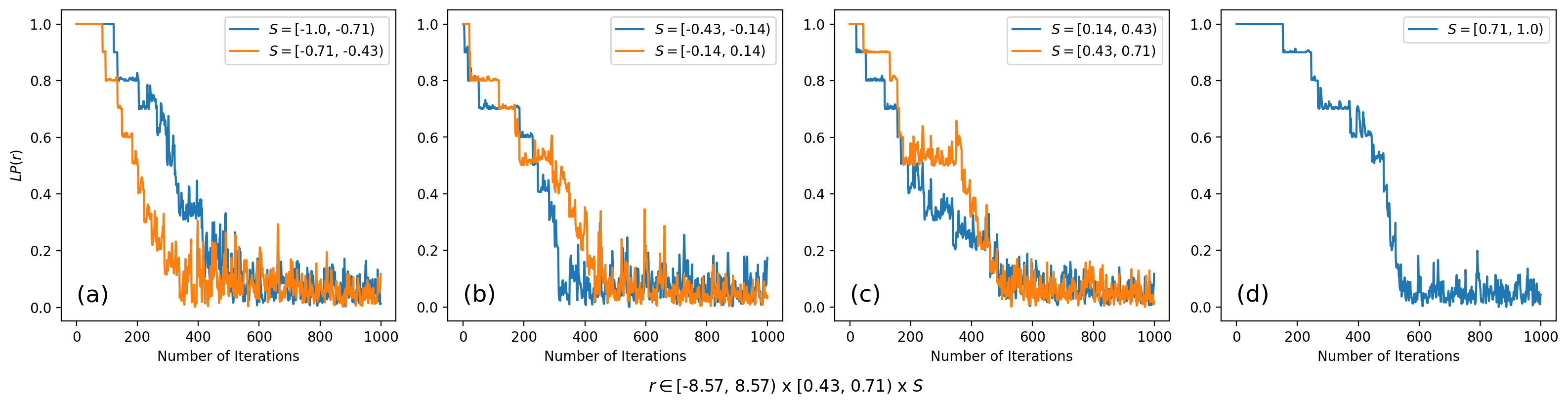}
    \caption{Typical LP profiles of illustrative regions defined with $\alpha, pos_x, pos_y \in [-8.57, 8.57] \times [-0.43, 0.71) \times S$ are shown, where $S$ is indicated in the legend of each LP curve. For ease of comparison some LP curves are superimposed.}
    \label{fig:singular_lp_values}
\end{figure}

\begin{figure}[!h]
    \centering
    \subfigure{ %
        \adjustbox{height=0.213\textheight}{%
          \includegraphics{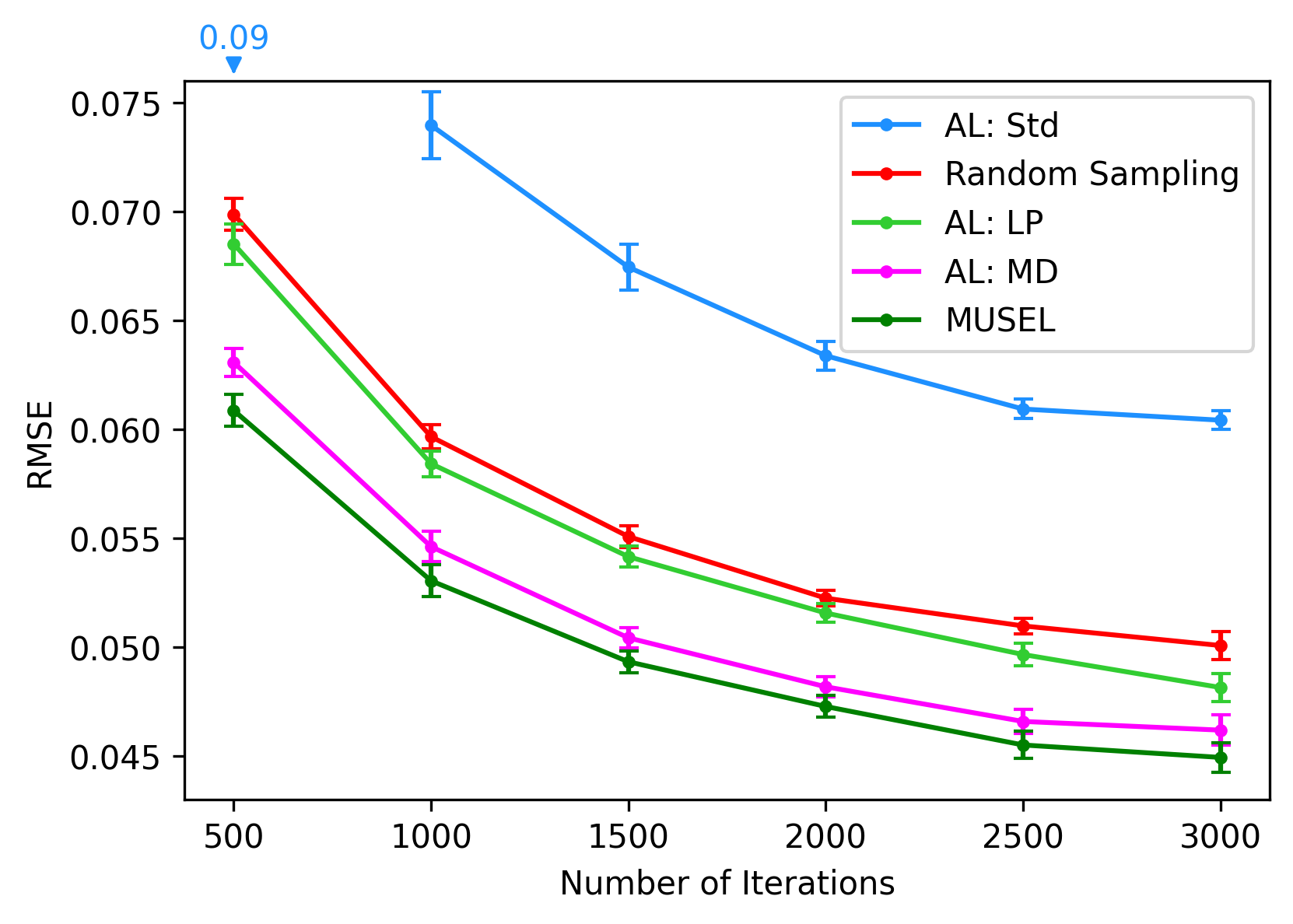}%
        }%
    }%
    \hfill
    \subfigure{%
        \adjustbox{height=0.2\textheight}{%
          \includegraphics{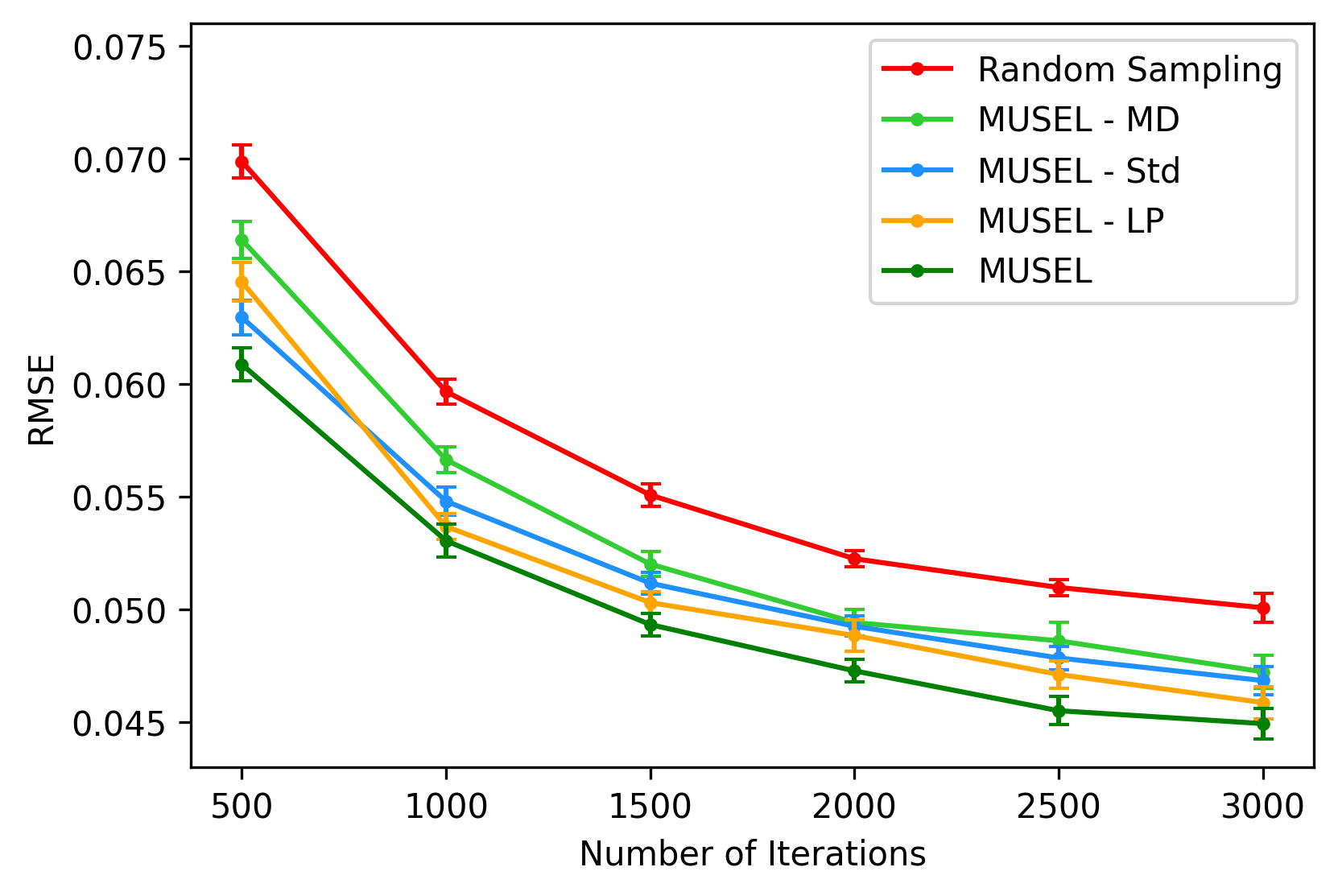}%
        }%
    }%
    \caption{Left: RMSE values obtained with single-measure-based and random sampling are contrasted with those of MUSEL as a function of sampling iterations. Error bars indicate the standard error of the mean (SEM). Right: Performance of MUSEL is contrasted with its performance when individual sampling components are ablated.}
    \label{fig:one-sphere_rmse}
    \label{fig:single_and_ablations}
\end{figure}

To analyze the relationship between LP and $U_{data}$, the evolution of LP values in representative regions of the one-sphere task is monitored using a sampling strategy based solely on LP (see Fig.\ref{fig:singular_lp_values}).
LP in boundary regions (blue curves in (a) and (d); orange in (c)) remains elevated longer than in central regions, indicating higher $U_{model}$. Furthermore, the observed general trend is that LP approaches zero by 500 iterations, matching the model convergence  (Fig.~\ref{fig:one-sphere_rmse}a).
These findings align with the assumption that lower LP values indicate a greater likelihood of irreducible error. It is worth noting, however, the high variability between consecutive values suggests that LP should be supplemented with other metrics for more accurate data uncertainty estimation, as discussed in Section~\ref{subsec:estimating_model_uncertainty}.

\subsection{One-Sphere Interaction Experiments}

To assess the efficacy of model uncertainty-based sampling in robot self-learning, two comparative tests were conducted. In each test, line 7 of Algorithm~\ref{algo:al} was replaced with the desired sampling strategy. Both evaluations included the full MUSEL framework as well as a random sampling baseline that draws $k=1$ i.i.d. samples from the input space.
The first evaluation compared model uncertainty estimation with each of its individual components used as the sole sampling criterion—these components are also state-of-the-art methods tested in finite input spaces, such as GSx \cite{wu2019active}. The second evaluation consisted of ablation experiments to quantify the impact of omitting each component from MUSEL.
\vspace{-4mm}
\vspace{-8mm}
\subsubsection{Sample Efficiency Comparison} \label{ssubsec:one_sphere_comp}
Figure~\ref{fig:one-sphere_rmse} summarizes the learning performance of all methods from both evaluations with RMSE values for each sampling method are compared at specific iteration intervals.
In the single‐measure sampling comparison (Figure~\ref{fig:one-sphere_rmse}, left), $\sigma$‐only selection yielded the lowest performance, which is the only one that random sampling can outperform. LP sampling marginally exceeded random selection, owing to its region‐level granularity. MD-based selection closely approached the performance of MUSEL, although MUSEL generally achieved a lower mean RMSE.

In the ablation experiments (Figure~\ref{fig:one-sphere_rmse}, right panel), as expected, discarding MD has the greatest effect on sample efficiency. However, surprisingly, the method still exhibits a better learning curve than when using $\sigma$- or LP-based selection alone, suggesting that these components produce a synergistic effect when combined. In contrast, removing the LP component results in the smallest reduction in performance, likely due to its regional selection strategy. Similarly, ablating the $\sigma$ component leads to a comparable decline in performance, possibly because the remaining criteria can still capture model uncertainty without explicitly incorporating $\sigma$.

\begin{figure}[!bhtp]
    \centering
    \subfigure{ 
        \adjustbox{height=0.19\textheight}{
          \includegraphics{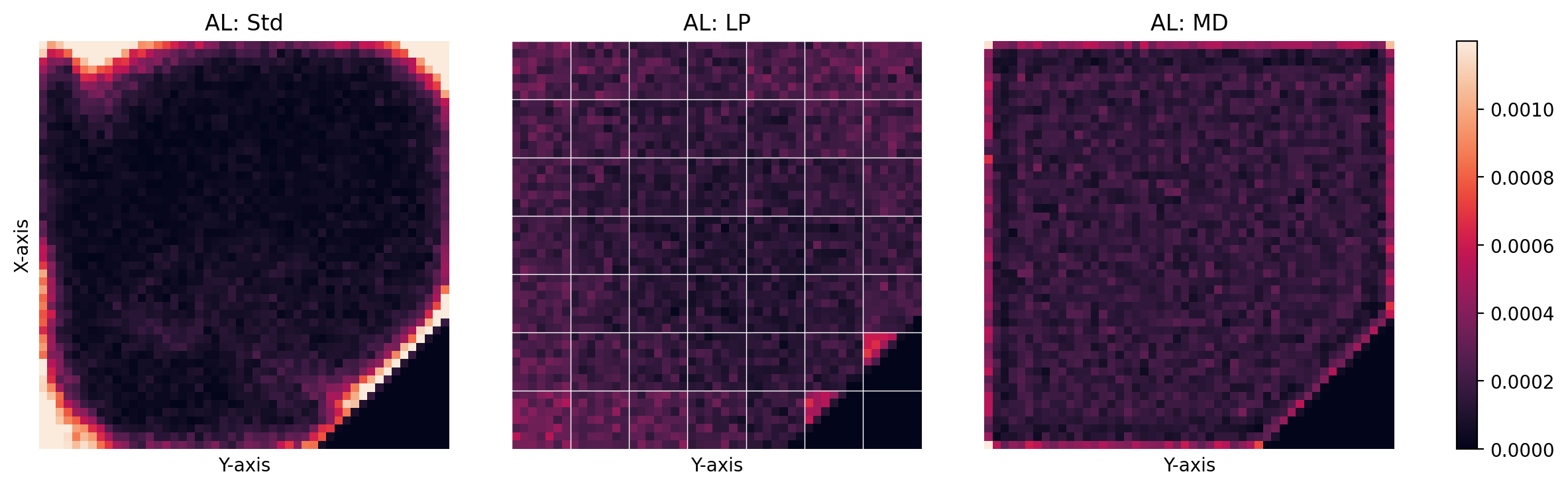}%
        }
        \label{subfig:ss_ind}
    }
    \subfigure{
        \adjustbox{height=0.19\textheight}{
          \includegraphics{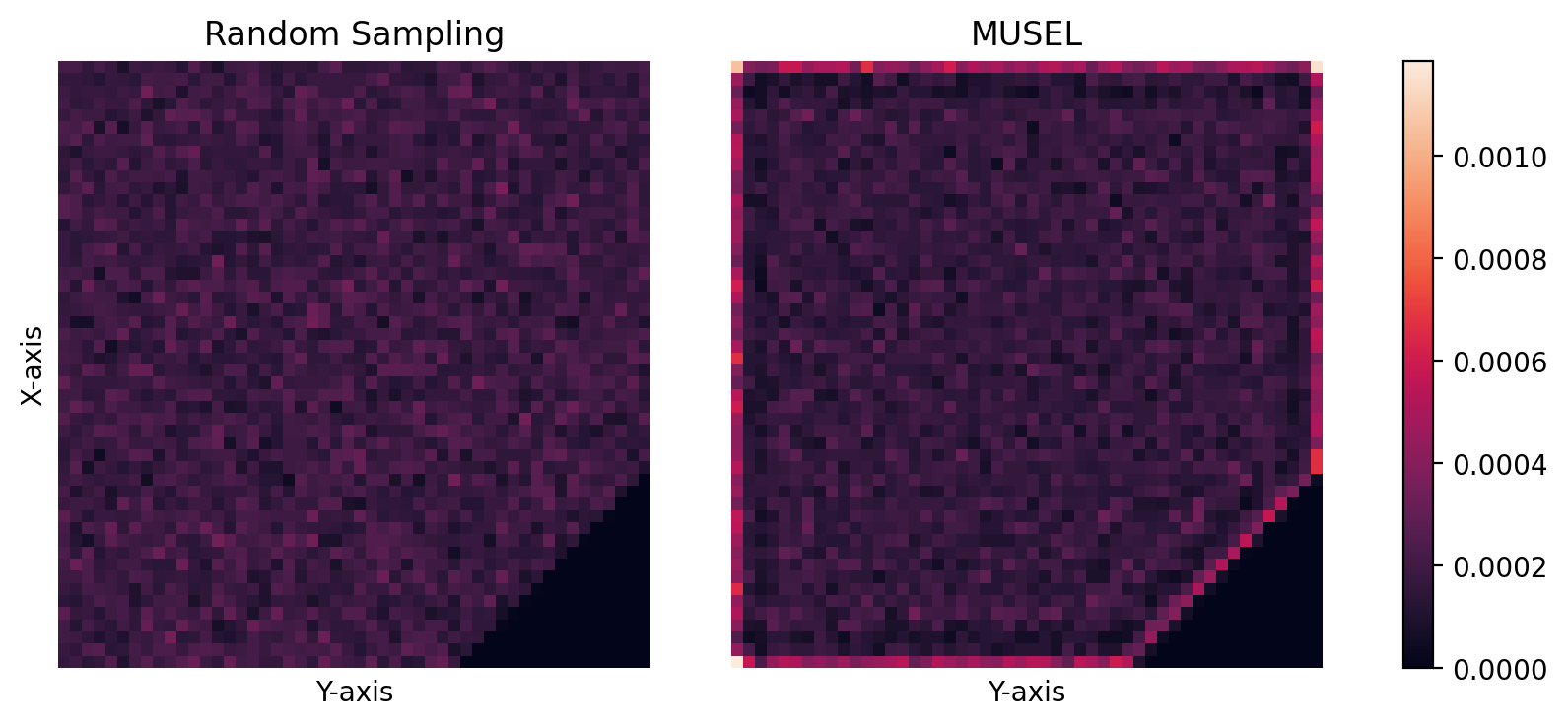}
        }%
        \label{subfig:ss_musel}
    }%
    \caption{Histogram of sample selection by random sampling (left) and MUSEL (right) over the space of $pos_x, pos_y$ with a grid of 50 x 50 over 10 repetitions. The histograms are normalized and scaled logarithmically for better visibility, and boundary of the regions in LP-based sampling are shown as white lines.}
    \label{fig:one-sphere_ss}
\end{figure}
\vspace{-5mm}
\subsubsection{Sample Selection Comparison}
To understand the observed learning performances, sampling histograms of the two methods over the ${pos_x, pos_y}$ space are shown in Figure~\ref{fig:one-sphere_ss}. Here, as expected, random sampling produces a uniform distribution, while other strategies concentrate data near the boundaries. This is reasonable, as the sphere’s behavior is more complex at the boundaries, requiring more samples to learn action-effect patterns.
$\sigma$-based sampling excessively targets these regions and fails to reduce $\sigma$ due to persistent data uncertainty, thereby limiting training input diversity and impeding generalization across the task.
The LP-based strategy, as it is limited to grid-based measurements and cannot differentiate inputs within a region, produces a broader yet still boundary-focused sampling pattern.
In contrast, both MD-based and MUSEL strategies follow a more balanced approach, as they can both detect data uncertainty and more effectively differentiate individual inputs. The MD-based strategy is agnostic to the performance of the learning engine, but it prioritizes boundary sampling based on the distribution of nearest inputs, which coincides with regions of higher model uncertainty in this robotic task.
\begin{table}[ht!]
\centering
\begin{tabular}{|l|c|c|c|c|c|c|}
\hline
\multirow{2}{*}{\textbf{Method}} 
  & \multicolumn{6}{|c|}{\textbf{Iterations}} \\ 
\cline{2-7}
  & \textbf{500} & \textbf{1000} & \textbf{1500} & \textbf{2000} & \textbf{2500} & \textbf{3000}\\
\hline
MUSEL &  104.2 & 189.5 & 266.2 & 342.3 & 418.0 & 487.3 \\ 
MD    & 117.8 & 221.0 & 311.4 & 398.0 & 478.2 & 562.5 \\ 
\hline
\end{tabular}
\vspace{0.2cm}
\caption{Counts for two methods across four iteration levels.}
\label{tab:boundary_counts}
\end{table}
\vspace{-5mm}

To further analyze the boundary focus of the methods, a boundary-count table was constructed to show the number of samples selected by each strategy (see Table~\ref{tab:boundary_counts}). As shown, MUSEL adopts a more balanced sampling approach, resulting in modest improvements in RMSE (see Figure~\ref{fig:one-sphere_rmse}a).

\subsection{Two-Sphere Interaction Experiments}

\subsubsection{Learning Efficiency Comparison}

In this section, we compare MUSEL to random sampling and MD-based sampling, the latter being the strongest competitor to model uncertainty strategy, according to the one-sphere task experiments. 

\begin{figure}
    \centering
    \includegraphics[height=0.21\textheight]{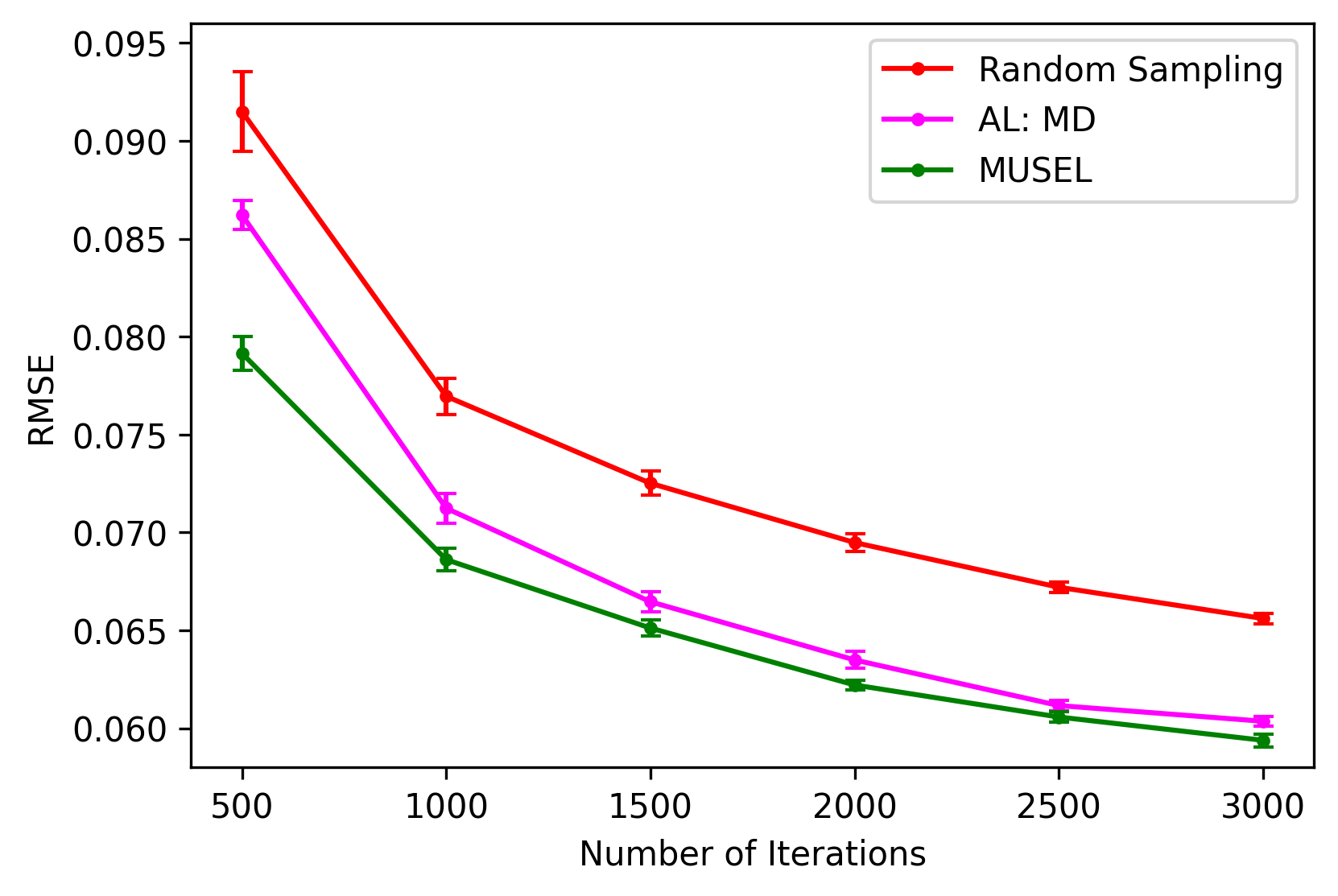}
    \caption{Comparison of MUSEL with MD-based and random sampling strategies is presented in the two-sphere interaction task. The test settings follows the same convention used in the one-sphere task (Figure~\ref{fig:one-sphere_rmse}).}
    \label{fig:two_sphere_comp}
\end{figure}
\vspace{-4mm}
The performance of the strategies is presented in Figure \ref{fig:two_sphere_comp} using the conventions established in the one-sphere interaction task. The results show that MUSEL and MD-based sampling achieve higher sample efficiency compared to random sampling, paralleling the findings from the one-sphere experiments. An interesting observation is that the improvement over random selection is even more pronounced in the two-sphere interaction task (see Figure \ref{fig:one-sphere_rmse}). 
This confirms the intuition that more complex learning tasks benefit more from guided sample selection. Although MD-based sampling--the closest competitor in the one-sphere task (see Section~\ref{ssubsec:one_sphere_comp})--outperforms random selection and remains competitive with MUSEL, the performance gap between the two methods is more pronounced in this setting, especially during the first half of training. These findings demonstrate the superior sample efficiency of MUSEL over MD-based sampling in more complex tasks.

\section{Conclusion}
\label{sec:conclusion}
In this study, we introduced MUSEL (Model Uncertainty for Sample-Efficient Learning), a novel sample-efficient active learning framework for robot self-supervised learning. The effectiveness of MUSEL is demonstrated in simulated robotic experiments.
A model uncertainty based sample selection strategy is also introduced to guide the framework for sample-efficient learning. This is achieved through the novel use of learning progress (LP) and minimum-distance (MD) metrics, enabling MUSEL to adaptively choose the most informative actions and address the high costs associated with extensive data collection in robot learning. In our experiments, when compared with the other baselines, our strategy achieves the lowest RMSE values during learning. Importantly in the more complex task, early convergence is observed with MUSEL, which further adds to its  suitability for robot learning tasks. 

Future work will involve testing the model in real-world robotic scenarios and addressing more complex tasks. Additionally, efforts will focus on improving the model’s computational efficiency and reducing the boundary biases introduced by the MD component to further enhance performance.

\section{ACKNOWLEDGMENTS}
This work was supported by JSPS Kakenhi Grant Number JP23K24926 and also JP25H01236.

\bibliographystyle{splncs04}
\bibliography{references}
\end{document}